# Efficient Decomposed Learning for Structured Prediction

Rajhans Samdani                                                                                      RSAMDAN2@ILLINOIS.EDU
Dan Roth                                                                                                  DANR@ILLINOIS.EDU

## Abstract

Structured prediction is the cornerstone of several machine learning applications. Unfortunately, in structured prediction settings with expressive inter-variable interactions, exact inference-based learning algorithms, e.g. Structural SVM, are often intractable. We present a new way, Decomposed Learning (DecL), which performs efficient learning by restricting the inference step to a limited part of the structured spaces. We provide characterizations based on the structure, target parameters, and gold labels, under which DecL is equivalent to exact learning. We then show that in real world settings, where our theoretical assumptions may not completely hold, DecL-based algorithms are significantly more efficient and as accurate as exact learning.

## 1. Introduction

Structured output spaces occur in many machine learning applications which aim to label certain sets of interdependent variables where the dependencies between variables dictate what assignments are possible. Several techniques have been proposed for learning in structured prediction (Collins, 2002; Tsochantaridis et al., 2004; Taskar et al., 2004). Typical discriminative structural learning algorithms (e.g. Collins (2002); Tsochantaridis et al. (2004)) perform a *global* MAP inference over the **entire** (hence 'global') output space as an intermediate step. We refer to such learning techniques as global learning (GL). Global inference, and hence GL, can be slow for models with high-order, expressive relations between the output variables.

GL algorithms perform **exact** MAP inference as a black box which may be an overkill for several problems, making learning slow. To alleviate this, we propose a novel algorithm which restricts the MAP inference to a smaller part of the output space by using additional information about a) the actual gold labels, b) the constraints on the output space, and c) the underlying parameters, which we want to learn. Consequently, our algorithm is much more efficient than GL. We call our approach *Decomposed Learning* (DecL) as we decompose the inference into smaller inference procedures over more tractable output spaces. We prove that in some settings, DecL is guaranteed to be equivalent to GL. We present experiments in real-world settings (where our theoretical assumptions may not hold) and show that DecL, with small-sized and problem-specific decompositions, perform as well as GL, while being significantly faster.

Several existing works perform approximate inference during supervised structured learning to the same end. Such approaches can broadly be divided into those that relax the expressive interactions between output variables (Roth & Yih, 2005; Punyakanok et al., 2005; Sutton & Mccallum, 2009) during learning and those that relax the integrality constraints on assignments (Kulesza & Pereira, 2008; Finley & Joachims, 2008; Martins et al., 2009; Meshi et al., 2010). Some of the MCMC-based contrastive techniques (Hinton, 2002; Wick et al., 2011) are conceptually similar to DecL in that they use approximate gradient steps for learning. Our work is also related in spirit to Meshi et al. (2010) who consider a Linear Programming relaxation of the entire inference and perform parameter updates after small message-passing inference steps. However, unlike these techniques, we don't replace exact inference by approximate inference; instead, we perform *exact* inference on a *smaller* output space. The closest works to DecL are Pseudolikelihood-based techniques (Besag, 1974; Sontag et al., 2010) to learning; however, while Pseudolikelihood is consistent asymptotically, DecL aims to achieve equivalence to GL with a finite amount of data.

The outline of this paper is as follows. Sec. 2 introduces the problem and notation and Sec. 3 assays two extreme styles of structured prediction. We introduce our approach in Sec. 4 and provide theoretical results in Sec. 5. We finally present empirical results in Sec. 6.

## 2. Problem Setting

Consider a structured prediction setting where a $d$-dimensional input $\mathbf{x}$ is drawn from a space $\mathcal{X}$ and the output variable $\mathbf{y}$ is, w.l.o.g., a vector of binary labels $\{y_1, \ldots, y_n\}$ drawn from $\mathcal{Y} \in \{0,1\}^n$. The space $\mathcal{Y}$ may be specified by a set of declarative constraints which can be viewed as a form of specifying some domain knowledge over $\mathbf{y}$.

**Inference:** The labels in $\mathbf{y}$ are correlated and so it is advantageous to predict them simultaneously. As is typical, we express the prediction over all variables in $\mathbf{y}$ using a scoring function $f(\mathbf{x}, \mathbf{y}; \mathbf{w}) = \mathbf{w} \cdot \phi(\mathbf{x}, \mathbf{y})$ as

$$\arg\max_{\mathbf{y} \in \mathcal{Y}} f(\mathbf{x}, \mathbf{y}; \mathbf{w}) = \arg\max_{\mathbf{y} \in \mathcal{Y}} \mathbf{w} \cdot \phi(\mathbf{x}, \mathbf{y}), \quad (1)$$

where $\phi(\mathbf{x}, \mathbf{y}) \in \mathbb{R}^d$ are feature expressed over both $\mathbf{x}$ and $\mathbf{y}$, and $\mathbf{w} \in \mathbb{R}^d$ are weight parameters. We refer to the arg max inference above as MAP inference[1].

**Structural Learning and evaluation:** The focus of this work is on learning the weight parameter, $\mathbf{w}$, from a given collection of labeled training instances $D = (\mathbf{x^1}, \mathbf{y^1}), \ldots, (\mathbf{x^m}, \mathbf{y^m})$. As is standard, the quality of a learned hypothesis is measured using a loss function $\Delta : \{0,1\}^n \times \{0,1\}^n \to \mathbb{R}_{\geq 0}$, satisfying $\Delta(\mathbf{y}, \mathbf{y}) = 0, \forall \mathbf{y} \in \{0,1\}^n$.

We focus on two popular classes of scoring functions $f(\mathbf{x}, \mathbf{y}; \mathbf{w})$:

- **Singleton with constraints:** $f(\mathbf{x}, \mathbf{y}; \mathbf{w})$ is a sum of linear classifiers, $f_i(\mathbf{x})$, for individual $y_i$: $f(\mathbf{x}, \mathbf{y}; \mathbf{w}) = \sum_{i=1}^{n} y_i f_i(\mathbf{x}) = \sum_{i=1}^{n} y_i \mathbf{w_i} \cdot \mathbf{x}$. The variables contained in $\mathbf{y}$ interact solely via mutual constraints. The region of allowed outputs, $\mathcal{Y}$, is specified by these constraints. This model has been used in numerous applications, especially in Natural Language Processing (NLP) where sometimes the constraints are inherent to the problem e.g. tree constraints in dependency parsing (Koo et al., 2010) and sometimes they are added declaratively (Roth & Yih, 2005; Clarke & Lapata, 2006; Barzilay & Lapata, 2006; Roth & Yih, 2007; Clarke & Lapata, 2008; Choi & Cardie, 2009; Ganchev et al., 2010). Exact MAP inference with expressive constraints is often formulated using expensive Integer Linear Programming (ILP) techniques.

- **Pairwise Markov Networks:** For a Pairwise Markov Network (PMN), $f$ is defined over a graph with $n$ nodes and a set of edges given by $E$. In particular, $f$ is a sum of individual and pairwise potential functions, $\phi$, corresponding to nodes and edges of the graph: $f(\mathbf{x}, \mathbf{y}; \mathbf{w}) = \sum_{i=1}^{n} \phi_i(y_i, \mathbf{x}; \mathbf{w}) + \sum_{i,k \in E} \phi_{i,k}(y_i, y_k, \mathbf{x}; \mathbf{w})$. $f$ is linear in $\mathbf{w}$. While PMNs are typical to probabilistic graphical model settings (e.g. HMM and CRF (Lafferty et al., 2001)), in this paper, we consider PMNs in a max-margin setting *a la* Taskar et al. (2004). PMNs are used extensively in many structured prediction applications in computer vision (Boykov et al., 1998), computational biology (Meshi et al., 2010), NLP, and information extraction (Lafferty et al., 2001; Sarawagi & Cohen, 2004). We also consider the case when higher order declarative constraints are added on top of a PMN scoring function (Roth & Yih, 2005).

## 3. Structured Prediction: Learning

This section discusses two styles of learning the parameter $\mathbf{w}$ from the training data $D$: global learning and local learning, with their shortcomings.

**Global Learning** Given the inference procedure in (1) and training data $D$, a popular discriminative learning approach (Tsochantaridis et al., 2004; Taskar et al., 2004) is to minimize an SVM-style convex upper bound on the loss[2] over the training data:

$$l(w) = \sum_{j=1}^{m} \max_{\mathbf{y} \in \mathcal{Y}} (f(\mathbf{x^j}, \mathbf{y}; \mathbf{w}) - f(\mathbf{x^j}, \mathbf{y^j}; \mathbf{w}) + \Delta(\mathbf{y^j}, \mathbf{y})) \quad (2)$$

The inference step in (2), involving max, is performed globally over all the labels of $\mathbf{y}$ and hence we call this style, Global Learning (GL). GL tends to be slow which hinders applications with a large output space or a large number of training examples.

**Local Learning** For faster learning, several approximations to GL have been used which ignore certain structural interactions so that the rest of the structure is easier to learn. We call this general paradigm of learning by relaxing to a more local or easy-to-learn structure, Local Learning (LL). For instance, when highly expressive constraints are used over the structure, then dropping such constraints makes the structure more "local" and faster to learn: for singleton functions (Punyakanok et al., 2005; Barzilay & Lapata, 2006), ignoring constraints reduces the problem to learning $n$ independent binary classifiers $\mathbf{w_i}$; in case of sequential or tree-structured problems, the task reduces to learning with dynamic programming inference (Koo et al., 2010; Roth & Yih, 2005). In case of multi-label classification, ignoring interactions between labels reduces the problem to learning a binary classifier for each label. In most LL scenarios, the ignored constraints, if any, are injected back during inference. Refer to Punyakanok et al. (2005) for a detailed analysis and comparison of GL and LL for singleton scoring functions with constraints.

---

[1] While MAP is used to refer to probabilistic inference, we abuse the terminology here to convey similar import.

[2] Throughout this paper, we omit the usual $l_2$ regularization term for the sake of brevity.

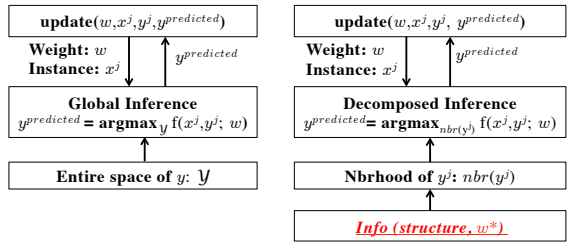

*Figure 1.* Figure highlighting the differences between typical GL and DecL. The diagrams on left and right illustrate the general scheme for GL and DecL, respectively. We show how DecL restricts learning-time inference by adding more information. During testing, we use exact inference.

LL schemes are much faster than GL; in general, however, LL fails to take advantage of the structure of $\mathcal{Y}$ which is where Decomposed Learning comes in.

## 4. Decomposed Learning (DecL)

For a training instance $(\mathbf{x^j}, \mathbf{y^j}) \in D$, let $nbr(\mathbf{y^j}) \subseteq \mathcal{Y}$ be a subset of the output space defining a "neighborhood" around $\mathbf{y^j}$, which is referred to as the ground truth or the gold output. The key idea behind decomposed learning (DecL) is to learn $\mathbf{w}$ by discriminating the supervised label $\mathbf{y^j}$ from only all $\mathbf{y'} \in nbr(\mathbf{y^j})$ instead of all $\mathbf{y'} \in \mathcal{Y}$. $nbr(\mathbf{y^j})$ can use additional information about the structure ($\mathcal{Y}$) and parameters which we intend to learn ($\mathbf{w}^*$) such that it captures the structure of $\mathcal{Y}$ while being much smaller. Fig. 1 shows the general schema for both GL and DecL, showing the similarities and the differences.

Let $N = \{nbr(\mathbf{y^j}) | j = 1, \ldots, m\}$ be the collection of neighborhoods for all training instances. To pursue the general idea behind our approach, we use a max-margin formulation (Taskar et al., 2004) for learning over given data $D = \{(\mathbf{x^j}, \mathbf{y^j})\}_{j=1}^m$. Specifically we minimize a loss function $DecL(\mathbf{w}; D, N)$ given by

$$\sum_{j=1}^m \max_{\mathbf{y} \in nbr(\mathbf{y^j})} \left( f(\mathbf{x^j}, \mathbf{y}; w) - f(\mathbf{x^j}, \mathbf{y^j}; w) + \Delta(\mathbf{y}, \mathbf{y^j}) \right). \quad (3)$$

The idea of looking at a smaller output space is natural; the key question is how do we create these neighborhoods so that the resulting learning algorithm is correct or at least gives a good approximation to GL. To motivate our technique for doing so, we use a simple example of multi-class classification with labels $1, \ldots, r$. This problem can be expressed as a structured prediction problem over $r$ binary variables $y_1, \ldots, y_r$ such that an instance with label $q$ is represented as a binary vector $\mathbf{y}[\mathbf{q}]$ obeying the constraint $\sum_i y[q]_i = 1$ and with $y[q]_i = 1$. For a training instance $(\mathbf{x}, \mathbf{y}[\mathbf{q}])$, GL aims to learn a scoring function which gives a score less than $\mathbf{y}[\mathbf{q}]$ to all other possible outputs $\mathbf{y}[\mathbf{i}]$, $\forall i \neq q$. Since the outputs are constrained such that any two outputs $\mathbf{y}[\mathbf{q}]$ and $\mathbf{y}[\mathbf{i}]$ differ on just the bits $y_q$ and $y_i$, this is achieved by merely comparing assignments to pairs of bits $y_q$ and $y_i$, $\forall i \neq q$. This is exactly what techniques like multi-class SVM (Crammer & Singer, 2002) and constrained classification (Har-Peled et al., 2003) do[3]. Overall, while multi-class classification is indeed a simple case as the space $\mathcal{Y}$ contains just $n$ outputs, we generalize the idea of creating neighborhoods over a large number of variables via smaller and more local comparisons.

We generate $nbr(\mathbf{y^j})$, by fixing a subset of the output variables to their gold labels in $\mathbf{y^j}$, while allowing the rest of them to vary, and repeating the same for different subsets of output variables. We formalize this idea through what we define as **decompositions** (hence, *decomposed* learning.) We give theoretically desirable properties of these neighborhoods in Sec. 5.

**Definition 1.** *Given a set of $n$ binary output variables indexed by $\{1, \ldots, n\}$, a **decomposition** $\mathcal{S}$ is a set containing distinct and non-inclusive (possibly overlapping) index sets which are subsets of $\{1, \ldots, n\}$: $\mathcal{S} = \{s_1, \ldots, s_l | \forall i, s_i \subseteq \{1, \ldots, n\}; \forall i, k, s_i \not\subset s_k\}$.*

Before explaining learning with decompositions, we give some notation. Given two output instances $\mathbf{y}, \mathbf{y'} \in \mathcal{Y}$, let $s(\mathbf{y}, \mathbf{y'})$ be the set indexing the differences between $\mathbf{y}$ and $\mathbf{y'}$ i.e. $s(\mathbf{y}, \mathbf{y'}) = \{i : y_i \neq y'_i\}$. Given a set $s \subseteq \{1, \ldots, n\}$, denote $-s = \{1, \ldots, n\} \setminus s$. Let $\mathbf{y_s} \in \{0, 1\}^{|s|}$ denote an assignment to the variables indexed by set $s$. Let $(\mathbf{y_s}, \mathbf{y^j_{-s}})$ be the output formed by replacing variables in $\mathbf{y^j}$ indexed by $s$ by corresponding variables in $\mathbf{y_s}$.

We associate one decomposition $\mathcal{S}^j$ with each training instance $(\mathbf{x^j}, \mathbf{y^j})$ and do inference during learning as follows. Given a gold output variable $\mathbf{y^j}$ pick a set $s \in \mathcal{S}^j$, fix variables in $\mathbf{y^j_{-s}}$ and look at all assignments to $\mathbf{y_s}$ such that $(\mathbf{y_s}, \mathbf{y^j_{-s}})$ **is feasible** (i.e. $\in \mathcal{Y}$); select the highest scoring assignment over all feasible selections of $\mathbf{y_s}$ and over all $s \in \mathcal{S}^j$ and return the structure. Given a decomposition $\mathcal{S}^j$ for $\mathbf{y^j}$, let the corresponding neighborhood be $nbr(\mathbf{y^j}, \mathcal{S}^j)$ given by $nbr(\mathbf{y^j}, \mathcal{S}^j) =$

---
[3]Interestingly, the one-vs-all technique ignores the given constraint (one-vs-all is a kind of LL technique) and may not be able to obtain linear separation even if the labels are pairwise linearly separable (Har-Peled et al., 2003).

**Algorithm 1** Subgradient-descent Alg. for DecL

1: **Given:** training data: $D = (\mathbf{x^j}, \mathbf{y^j})_{j=1}^m$; step sizes $\eta_t$
   decompositions: $S = (\mathcal{S}^1, \ldots, \mathcal{S}^m)$.
2: $\mathbf{w} \leftarrow \mathbf{0}$
3: **for** $t = 0$ to $T$ **do**
4:   **for** $j = 1$ to $m$ **do**
5:     $\mathbf{y}' \leftarrow \arg\max_{s \in \mathcal{S}^j, \mathbf{y_s} \in \{0,1\}^{|s|}:(\mathbf{y_s}, \mathbf{y^j_{-s}}) \in \mathcal{Y}}$
        $(f(\mathbf{x^j}, (\mathbf{y_s}, \mathbf{y^j_{-s}}); \mathbf{w}) + \Delta(\mathbf{y_s}, \mathbf{y^j_{-s}}))$
6:     $\mathbf{w} \leftarrow \mathbf{w} + \eta_t \left( \phi(x^j, \mathbf{y^j}) - \phi(x^j, \mathbf{y'}) \right)$
7:   **end for**
8: **end for**

$\{\mathbf{y} \in \mathcal{Y} | \exists s \in \mathcal{S}^j, \ s(\mathbf{y^j}, \mathbf{y}) \subseteq s\}$. Using the above style of inference results in minimizing the following convex function for learning

$$\begin{aligned}DecL(\mathbf{w}; D) &= \sum_{j=1}^m \max_{s \in \mathcal{S}^j, \mathbf{y_s} \in \{0,1\}^{|s|}:(\mathbf{y_s}, \mathbf{y^j_{-s}}) \in \mathcal{Y}} \\ (f(\mathbf{x^j}, (\mathbf{y_s}, \mathbf{y^j_{-s}}); \mathbf{w}) &- f(\mathbf{x^j}, \mathbf{y^j}; w) + \Delta(\mathbf{y_s}, \mathbf{y^j_{-s}})) \ .\end{aligned} \quad (4)$$

To minimize Eq. 4, we use a subgradient descent scheme shown in Alg. 1[4].

Let DecL-$k$ be the special case where all subsets of $\{1, \ldots, n\}$ of size $k$ ($k \geq 1$) are considered in the decomposition. For multi-class classification, DecL-2 with $\Delta$ as the Hamming loss is the same as multi-class SVM (Crammer & Singer, 2002) and Alg. 1 with DecL-2 and $\Delta = 0$ (perceptron loss) yields constrained classification (Har-Peled et al., 2003) thus closing our loop on multi-class classification. Note that, in Step 5 of Alg. 1, going over all sets in $\mathcal{S}^j$ to find arg max can be slow if the number of sets inside each decomposition is large (e.g. in DecL-$k$ for large $k$.) To get around this, we compute max over a few sets selected uniformly at random from the decomposition. One can also use more complicated convex optimization techniques which require evaluating the max over just one set at a time (Gaudioso et al., 2006).

In practice, decompositions can be guided by domain knowledge — highly coupled output variables should be put in the same set while somewhat unrelated variables should be kept separate. The complexity of learning is small if the sizes of the sets considered in the decomposition are small. Sec. 5 provides theoretical results on decompositions for certain cases.

## 5. Theoretical Analysis

Our theoretical anaylsis carries a different flavor than standard generalization bounds. We present theoretical results to show some conditions under which DecL

---

[4]Instead of subgradient-descent, DecL can also be used in a cutting-plane method (Tsochantaridis et al., 2004).

---

is equivalent to GL. We start with the trivial observation that when each neighborhood is equal to $\mathcal{Y}$, then DecL is the same as GL. Due to the lack of space, we have moved all the proofs to the supplement.

We assume that the data is separable. Our interest is in all parameters $\mathbf{w}^*$ which satisfy the following margin-separation condition $\forall (\mathbf{x^j}, \mathbf{y^j}) \in D$:

$$f(\mathbf{x^j}, \mathbf{y^j}; \mathbf{w}^*) \geq f(\mathbf{x^j}, \mathbf{y}; \mathbf{w}^*) + \Delta(\mathbf{y^j}, \mathbf{y}), \ \forall \mathbf{y} \in \mathcal{Y} \quad (5)$$

the set of which can be written (omitting regularization and using (2)) as $W^* = \{\mathbf{w} | l(\mathbf{w}) = 0\} \subseteq \mathbb{R}^d$. Let $W^{dec} = \{\mathbf{w} | DecL(\mathbf{w}; D, N) = 0\} \subseteq \mathbb{R}^d$ be the set of weights obtained by DecL[5] (we leave the neighborhoods selected for DecL implicit here.) Throughout this section, we assume that there exists at least one separating weight vector in $W^*$.

**Assumption 1:** $W^*$ is non-empty.

We use the following property to express our results.

**Exactness:** DecL is said to be *exact* if $W^{dec} = W^*$ for the given data $D$.

Our goal is to find small neighborhoods for DecL for which exactness holds. Note that the Pseudolikelihood-based approaches (Besag, 1974; Dillon & Lebanon, 2010; Sontag et al., 2010) to structured prediction are asymptotically consistent; that is, they are equivalent to GL only in the limit of infinite data. In practice, one uses a finite amount of data to obtain a weight vector by minimizing a convex regularizer on $\mathbf{w}$ (e.g. $\min \|\mathbf{w}\|_p$ for $p \geq 1$) while requiring separation (Cond. (5).) In this case, exactness, i.e. $W^{dec} = W^*$, implies that DecL and GL minimize the same regularization function over two equal sets — if the regularizer is strictly convex, they will output the same weight. Thus exactness is clearly a stronger and more useful property than asymptotic consistency. Our goal is to determine families of decompositions that will result in the exactness of DecL.

To analyze exactness of DecL, we use the following property to characterize the loss function $\Delta$.

**Subadditivity:** $\Delta(\mathbf{y}, \mathbf{y}')$ is subadditive if $\forall \mathbf{y}, \mathbf{y}', \mathbf{y^1}, \mathbf{y^2} \in \mathcal{Y}$, with $s(\mathbf{y}, \mathbf{y^1}) \cup s(\mathbf{y}, \mathbf{y^2}) = s(\mathbf{y}, \mathbf{y}')$, we have $\Delta(\mathbf{y}, \mathbf{y}') \leq \Delta(\mathbf{y}, \mathbf{y^1}) + \Delta(\mathbf{y}, \mathbf{y^2})$.

Several common loss functions like *Perceptron loss* i.e. no margin requirement, Hamming loss, and zero-one loss are subadditive. We now make the following simple observations.

**Observation 1 (Closed and Convex).** $W^*$ *is an*

---

[5]$W^*$ (and $W^{dec}$) is clearly not a singleton set as $l(\mathbf{w}^*) = 0 \Rightarrow l(\lambda \mathbf{w}^*) = 0 \ \forall \lambda \geq 1$.

*intersection of closed half spaces — one for each separation constraint given by (5). Thus $W^*$ is closed and convex. Similarly, $W^{dec}$ is closed and convex.*

**Observation 2 (Outer bound).** *For all decompositions, the set of separating weights for DecL give an outer-bound on the set of separating weights for GL, i.e. $W^* \subseteq W^{dec}$ as DecL seeks to separate the gold output from only a subset of the output space.*

Due to observation 2, to show that DecL is exact for some decompositions, we need only show that for any $\mathbf{w}' \notin W^*$, we also have $\mathbf{w}' \notin W^{dec}$ — since both $W^*$ and $W^{dec}$ are closed and convex, we need to show this only for $\mathbf{w}'$ immediately outside the boundary of $W^*$ (see the proof in the supplement.) To this end, we define $B(\mathbf{w}, \epsilon) = \{\mathbf{w}' | \ \|\mathbf{w}' - \mathbf{w}\| \leq \epsilon\}$ as a closed ball of radius $\epsilon$ centered around $\mathbf{w}$.

**Theorem 1.** *DecL is exact if $\forall \mathbf{w} \in W^*, \exists \epsilon > 0$, such that $\forall \mathbf{w}' \in B(\mathbf{w}, \epsilon), \forall(\mathbf{x^j}, \mathbf{y^j}) \in D$ the following condition holds for $nbr(\mathbf{y^j})$: if $\exists \mathbf{y} \in \mathcal{Y}$ with $f(\mathbf{x^j}, \mathbf{y}; \mathbf{w}') + \Delta(\mathbf{y^j}, \mathbf{y}) > f(\mathbf{x^j}, \mathbf{y^j}; \mathbf{w}')$ then $\exists \mathbf{y}' \in nbr(\mathbf{y^j})$ with $f(\mathbf{x^j}, \mathbf{y}'; \mathbf{w}') + \Delta(\mathbf{y^j}, \mathbf{y}') > f(\mathbf{x^j}, \mathbf{y^j}; \mathbf{w}')$.*

This theorem essentially requires that a $\mathbf{w}'$ which does not globally separate examples in $D$, also does not separate the decomposed learning examples. We note that this theorem is very general and applies to any structured prediction problem (and any $\Delta$.) We use this theorem to prove exactness for certain decompositions based on some easy to determine characterizations of a) the structure ($\mathcal{Y}$), b) the correct parameters ($W^*$), and c) the data $D$. The following corollary is an immediate consequence of Theorem 1. Roughly, this corollary requires that the difference between the score of the gold output and that of any other output is bounded by the sum of score differences between the gold output and that of outputs in the neighborhood.

**Corollary 1.** *DecL is exact if $\Delta$ is subadditive and $\forall \mathbf{w} \in W^*, \exists \epsilon > 0$ such that $\forall \mathbf{w}' \in B(\mathbf{w}, \epsilon), \forall(\mathbf{x^j}, \mathbf{y^j}) \in D, \forall \mathbf{y} \in \mathcal{Y}, s(\mathbf{y}, \mathbf{y^j})$ can be partitioned into sets $s_1, \ldots, s_l$ such that $\forall k \in \{1, \ldots, l\}, (\mathbf{y_{s_k}}, \mathbf{y^j_{-s_k}}) \in nbr(\mathbf{y^j}, \mathcal{S}^j)$ and*

$$f(\mathbf{x^j}, \mathbf{y}; \mathbf{w}') - f(\mathbf{x^j}, \mathbf{y^j}; \mathbf{w}') \leq$$
$$\sum_{k=1}^{l} \left( f(\mathbf{x^j}, (\mathbf{y_{s_k}}, \mathbf{y^j_{-s_k}})); \mathbf{w}') - f(\mathbf{x^j}, \mathbf{y^j}; \mathbf{w}') \right) . \quad (6)$$

Using these general results, we now examine two different classes of scoring functions mentioned in Sec. 2.

### 5.1. Exactness of DecL for Singleton Scoring Functions with Constraints

In this section, we present exactness results for DecL with singleton scoring function $f(\mathbf{x}, \mathbf{y}; \mathbf{w}) = \sum_{i=1}^{n} y_i f_i(\mathbf{x}) = \sum_{i=1}^{n} y_i \mathbf{w_i} \cdot \mathbf{x}$ where the space $\mathcal{Y}$ is specified by constraints. For instance, $\mathcal{Y}$ can be specified by a collection of $l$ logical constraints: $\mathcal{Y} = \{\mathbf{y} \in \{0,1\}^n \mid C_k(\mathbf{y}) = 1, k = 1, \ldots, l\}$ where $C_k$ is a logical function (e.g. OR) over binary variables in $\mathbf{y}$. $\mathcal{Y}$ can also be specified by linear constraints over $\mathbf{y}$ as $\mathcal{Y} = \{\mathbf{y} \in \{0,1\}^n | A\mathbf{y} \leq \mathbf{b}\}$.

In several practical applications, the constraint structure has some symmetry to it and we can invoke Cor. 1 to provide exactness guarantees for decompositions with set sizes independent of the number of variables. The following corollaries apply to two such cases with set sizes only dependent on the number of constraints.

**Corollary 2.** *If $\mathcal{Y}$ is specified by $k$ `OR` constraints, then Decl-$(k+1)$ is exact for subadditive $\Delta$.*

**Corollary 3.** *If $\mathcal{Y}$ is specified by $k$ ($k \geq 1$) linear constraints: $A\mathbf{y} \leq b$ (or '$\geq$', '='), where $A$ is a binary matrix such that any two variables in $\mathbf{y}$ participate in at most one constraint, Decl-$3k$ is exact for subadditive $\Delta$.*

As a consequence of Cor. 2, if the space $\mathcal{Y}$ is specified by $k$ horn clauses (Srikumar & Roth, 2011), then DecL-$(k+1)$ is exact regardless of the number of variables in each clause.

We also note that the results in this section are based on worst-case analyses. In practice, much smaller-sized decompositions work well in most cases.

### 5.2. Exactness for Pairwise Markov Networks

While in the last section, we presented exactness results solely based on constraints, in this section, we present decompositions for binary PMNs using some knowledge about the true parameters $W^*$.

Recall that, for PMNs, $f(\mathbf{x}, \mathbf{y}; \mathbf{w}) = \sum_{i=1}^{n} \phi_i(y_i, \mathbf{x}; \mathbf{w}) + \sum_{i,k \in E} \phi_{i,k}(y_i, y_k, \mathbf{x}; \mathbf{w})$ where $E$ is the set of edges for the underlying graph. Inference over such functions is NP hard in general.

A pairwise potential function, $\phi_{i,k}$ is called *submodular* if $(\phi_{i,k}(1,1) + \phi_{i,k}(0,0)) - (\phi_{i,k}(1,0) + \phi_{i,k}(0,1)) > 0$ i.e. it prefers similar labels; it is called *supermodular* if $(\phi_{i,k}(1,1) + \phi_{i,k}(0,0)) - (\phi_{i,k}(1,0) + \phi_{i,k}(0,1)) < 0$.

**Assumption 2:** Assume that $\forall (i,k) \in E, \forall \mathbf{x^j} \in D, \forall \mathbf{w}^* \in W^*, \phi_{ik}(\cdot, \cdot, \mathbf{x^j}; \mathbf{w}^*)$ is either submodular or supermodular; also, we know if any given $\phi_{ik}$ is submodular or supermodular.

Such knowledge about pairwise potential functions is often available in practice, especially for submodular potentials. For instance, in several computer vision tasks, neighboring pixels are more likely to carry the same label (Besag, 1986; Boykov et al., 1998); in infor-mation extraction tasks, consecutive words are likely to belong to the same field. We can also approximately determine this information by computing mutual information over labeled data. With Assumption 2, we present a decomposition, which leads to exactness.

For each instance $(\mathbf{x}, \mathbf{y^j} = \{y_1^j, \ldots, y_n^j\})$, define $E^j = \{(u,v) \in E | (\phi_{uv} \text{ is submodular and } y_u^j = y_v^j) \text{ or } (\phi_{uv} \text{ is supermodular and } y_u^j \neq y_v^j)\}$ i.e. $E^j$ removes those edges from $E$ where the labels on nodes "disagree" with the corresponding pairwise potentials. With $(\mathbf{x^j}, \mathbf{y^j})$, we associate a decomposition $\mathcal{S}_{pair}(\mathbf{y^j}) = \{c_1, \ldots, c_l\}$ where $c_1, \ldots, c_l$ are indices of the maximal connected components in $E^j$.

**Theorem 2.** *For PMNs where Assumption 2 is satisfied, DecL with $\mathcal{S}_{pair}$ is exact with subadditive $\Delta$.*

As a simple illustration, consider a sequential HMM where it is known that the same-state transition probabilities are higher than those for different states i.e. all $\phi_{ik}$ are submodular. Then for $\mathbf{y^j} = 1110011$, the corresponding decomposition is $\mathcal{S}_{pair} = \{\{1,2,3\}, \{4,5\}, \{6,7\}\}$ as it contains the maximal connected components with the same label.

Notably, graph cuts can be used for efficient learning over binary PMNs with submodular potentials (Szummer et al., 2008). We note that with submodular potentials, we can augment decomposed inference in DecL with graph cuts in a similar fashion to make it even more efficient.

Finally, DecL can be used when certain additional global constraints are added to PMNs (Roth & Yih, 2005). The exactness guarantees hold for $\mathcal{S}_{pair}(\mathbf{y^j})$ if $\forall \mathbf{y} \in \mathcal{Y}, \forall s \in \mathcal{S}_{pair}(\mathbf{y^j}), (\mathbf{y}_s, \mathbf{y}^\mathbf{j}_{-s}) \in \mathcal{Y}$. Exact global inference can replace DecL inference for those training examples where this condition does not hold. In practice, we find (see Sec. 6.3) that $\mathcal{S}_{pair}$ works very well for non-binary PMNs in the presence of constraints, where some of our assumptions do not hold.

## 6. Experiments

This section presents experimental results on non-ideal real world settings showing that DecL is effective and robust. We show results on synthetic data as well as two real-world tasks of multi-label classification and information extraction. We perform exact inference using ILP, wherever needed. We show that DecL performs favorably relative to GL on these tasks while greatly reducing the training time. We use appropriate LL approaches as competing baselines. In settings with constraints, we consider another baseline, LL+C, that uses constraints — which were ignored during learning — for inference during testing.

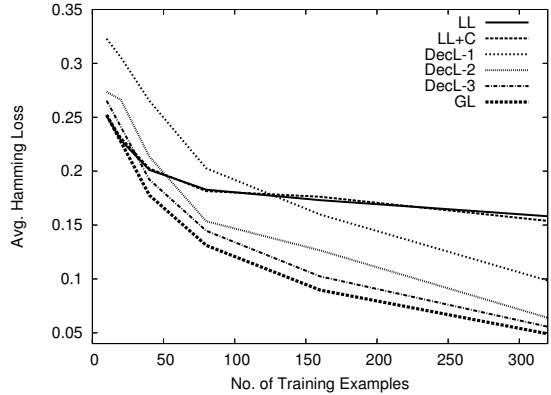

*Figure 2.* Synthetic data: Avg. Hamming loss for different learning algorithms. LL: local learning; LL+C: local learning with constraints during inference; GL: global learning; DecL-1,2,3: our approach. Note that LL and LL+C are really close. LL+C fails to obtain low error even with a large amount of training data, while Decomposed Learning algorithms achieve a continuously diminishing error.

*Table 1.* Synthetic Data: Avg. training times in seconds for different learning algorithms for training size=320.

| LL | DecL-1 | DecL-2 | DecL-3 | GL |
|---|---|---|---|---|
| 1.7 | 12.33 | 29.8 | 44.25 | 1181.47 |

### 6.1. Experiments on Synthetic Data

We first analyze DecL with simple decompositions — DecL-1,2,3 — with singleton scoring functions in a controlled synthetic setting by measuring the performance and efficiency. We generate data with 10 binary output variables, which are constrained by randomly generated linear constraints. We ensure that the resulting $\mathcal{Y}$ contains at least 50 outputs. The features $\mathbf{x}$ are sampled randomly from a 20-dimensional space. We randomly generate singleton scoring functions and determine gold labelings for each instance as per Eq. 1 (thus we know that the data is separable.)

For learning, we use SVM-Struct (Tsochantaridis et al., 2004) to implement our algorithms. Our LL baseline ignores the constraints during learning thereby reducing it to learning 10 independent binary classifiers. We test on 200 instances and tune $C$, the regularization penalty parameter, on 100 validation instances; we average over 10 random trials. Fig. 2 plots the loss for each algorithm against the size of the training data and Tab. 1 shows the training time on 320 examples. Note that the training time for DecL could be much lower with preprocessing of data.

We observe that although LL and LL+C exhibit relatively low error even with a small amount of data, they fail to converge to a near-perfect classifier like GL, with a large amount of data. On the other hand,

*Table 2.* Multi-label classification: Avg. F1 and training times for the Reuters data. We report total time spent on inference during learning. The differences in avg. F1 between DecL-2,3 and GL are not statistically significant.

| LL | DecL-1 | DecL-2 | DecL-3 | GL |
|---|---|---|---|---|
| **Avg. F1** (in % points) | | | | |
| 79.80 | 54.22 | **81.46** | **82.56** | **81.81** |
| **Time** (x 1000 seconds) | | | | |
| 8.08 | 5.09 | **22.86** | **68.10** | 126.78 |

DecL-2, 3 exhibit performance close to global learning while taking much less time to train.

### 6.2. Multi Label Document Classification

We test various algorithms on a multi-label document classification task over the Reuters dataset (Lewis et al., 2004). We use one section of the data with 6,000 instances and reduce it to the 30 most frequent labels. We keep 3600 instances for training, 1200 for testing, and 1200 for validation.

We model the scoring function as a PMN over a complete graph over all the labels to capture interactions between all pairs of labels. We compare DecL-1,2,3 with GL and a local learning (LL) approach which drops the pairwise components reducing the problem to 30 independent binary classifiers. We again use SVM-Struct for learning the parameters for GL and DecL. We measure the performance using a per-instance F1 measure given by $F1 = \frac{2c}{t+p}$, where $t$ is the number of gold labels for this instance, $p$ is the number of predicted labels, and $c$ is the number of correct predictions; we report averages over all test instances. Tab. 2 presents results with 10-fold cross validation[6]. DecL-2,3 perform as well as GL and much better than LL. Notably, DecL-2 is 6 times faster than GL. As in the synthetic data experiments, DecL-1, a.k.a. Pseudomax (Sontag et al., 2010), performs badly.

### 6.3. Information Extraction: Sequence Tagging with Submodular Potentials

We test the efficacy of our approach on two information extraction tasks inspired by our analysis of PMNs in Sec. 5.2. Our task is to identify the functional fields (e.g. 'author', 'title', 'facilties', 'roommates') from citations (McCallum et al., 2000) and advertisements (Grenager et al., 2005) datasets. We model this setting as an HMM (a special case of PMN) with different functional fields as hidden states and words as emissions. We add certain global constraints borrowed from Chang et al. (2007) to the HMM, which necessitate ILP-based inference.

---
[6]We observe similar results for averaged per-label F1.

*Table 3.* Performance comparison showing average accuracy (Acc) of HMM, LL+C, GL, and DecL. CRR07 refers to the state-of-the-art supervised results reported on these datasets by Chang et al. (2007). We also show average training time (Time) taken for GL and DecL in hours. Size indicates the number of documents.

| Size | HMM | LL+C | CRR07 | GL | | DecL | |
|---|---|---|---|---|---|---|---|
| | Acc | Acc | Acc | Acc | Time | Acc | Time |
| **Citations dataset** | | | | | | | |
| 20 | 66.74 | 77.00 | 81.1 | 79.60 | 7.97 | 79.26 | 1.05 |
| 300 | 91.21 | 91.52 | 92.5 | **94.55** | 40.59 | **94.77** | 10.69 |
| **Ads dataset** | | | | | | | |
| 20 | 67.78 | 71.57 | 71.9 | 69.05 | 22.76 | 69.18 | 11.57 |
| 100 | 76.52 | 78.97 | **80.4** | 80.0 | 75.46 | **80.3** | 37.55 |

For the given tasks, words corresponding to a field, e.g. 'title', occur in long contiguous blocks; thus we assume that the correct HMM transition matrix has a high same-state transition probability which is a generalization of the submodular potentials we assumed in Sec. 5.2 and hence a natural testing ground for our theory. We use the decomposition $\mathcal{S}_{pair}$ presented in Sec. 5.2 to perform DecL; intuitively, these decompositions enable DecL to capture the "diagonal-heavy" nature of the HMM transition matrix while allowing it to learn the transitions between different fields.

We perform discriminative learning using averaged structured perceptron (Collins, 2002). We use HMM without constraints as an LL baseline. We obtain an LL+C baseline by adding constraints during test.

Table 3 presents the results for the two domains. LL-based approaches perform very well for small data sizes because with a less expressive model, they need less data to generalize. However, with large amounts of training data, GL and DecL easily outperform HMM and LL+C. DecL does slightly, although not significantly, better than GL while being 2-8 times faster. Our results compare favorably with the state-of-the-art supervised results reported on these datasets by Chang et al. (2007) (CRR07.) Overall, we gather that by utilizing very simple knowledge of the task at hand (submodular potentials), we can perform near-global learning while being very efficient.

### 7. Conclusion

We presented Decomposed Learning (DecL) — a technique for efficient structural learning. DecL learns efficiently by performing inference over a small part of the output space. We provided theoretical results, which use characterizations of the structure, target parameters, and ground truth labels to decompose the output space such that the resulting DecL is efficient and equivalent to exact learning. While the common approximation practice in structural learning tasks is to use approximate MAP inference without guarantees, our approach may provide a way to achieve significant improvements in these cases and can be augmented with existing approximation techniques like LP-relaxation. Indeed, our experimental results indicate that our algorithms are robust and perform very well on real world data.

**Acknowledgments:** This research is sponsored by the Army Research Laboratory (ARL) under agreement W911NF-09-2-0053, Defense Advanced Research Projects Agency (DARPA) Machine Reading Program under Air Force Research Laboratory (AFRL) prime contract no. FA8750-09-C-018, and an ONR Award on Guiding Learning and Decision Making in the Presence of Multiple Forms of Information. Any opinions, findings, conclusions or recommendations are those of the authors and do not necessarily reflect the views of the funding agencies.